\documentclass[conference]{IEEEtran}
\IEEEoverridecommandlockouts
\usepackage{cite}
\usepackage{amsmath,amssymb,amsfonts}
\usepackage{algorithmic}
\usepackage{graphicx}
\usepackage{textcomp}
\usepackage{xcolor}
\def\BibTeX{{\rm B\kern-.05em{\sc i\kern-.025em b}\kern-.08em
    T\kern-.1667em\lower.7ex\hbox{E}\kern-.125emX}}

\usepackage{algorithm}
\usepackage{array}
\usepackage[caption=false,font=normalsize,labelfont=sf,textfont=sf]{subfig}
\usepackage{stfloats}
\usepackage{url}
\usepackage{verbatim}

\usepackage{multirow}
\usepackage{hyperref}
\hypersetup{
    colorlinks=true,
    linkcolor=blue,
    urlcolor=cyan,
}

\newcommand{\PreserveBackslash}[1]{\let\temp=\\#1\let\\=\temp}
\newcolumntype{C}[1]{>{\PreserveBackslash\centering}p{#1}}

\begin{document}

\title{SGM3D: Stereo Guided Monocular 3D Object Detection
\thanks{$\dag$ Li Zhang is the corresponding author at School of Data Science and Shanghai Key Laboratory of Intelligent Information Processing, Fudan University. Email: lizhangfd@fudan.edu.cn}
}

\author{Zheyuan Zhou$^{1} $ 
        \quad Liang Du$^{1}$
        \quad Xiaoqing Ye$^{2}$
        \quad Zhikang Zou$^2$ 
        \quad Xiao Tan$^2$ \\
        Li Zhang$^{1 \dag }$
        \quad Xiangyang Xue$^1$ 
        \quad Jianfeng Feng$^1$ 
        \vspace{0.8em} 
        \\
        
        $^1$ \textit{Fudan University}
        \quad \quad $^2$ \textit{Baidu Inc.} \\
}

\iffalse
\author{
\IEEEauthorblockN{Zheyuan Zhou\textsuperscript{1} }
\IEEEauthorblockA{\textit{Fudan University} \\
}
\and
\IEEEauthorblockN{Liang Du\textsuperscript{1} }
\IEEEauthorblockA{\textit{Fudan University} \\
}
\and
\IEEEauthorblockN{Xiaoqing Ye\textsuperscript{2} }
\IEEEauthorblockA{\textit{Baidu Inc.} \\
}
\and
\IEEEauthorblockN{Zhikang Zou\textsuperscript{2} }
\IEEEauthorblockA{\textit{Baidu Inc.} \\
}
\and
\IEEEauthorblockN{Xiao Tan\textsuperscript{2} } 
\IEEEauthorblockA{\textit{Baidu Inc.} \\
}
\and
\IEEEauthorblockN{Errui Ding\textsuperscript{2} }
\IEEEauthorblockA{\textit{Baidu Inc.} \\
}
\and
\IEEEauthorblockN{Li Zhang\textsuperscript{1}$^{\dag}$}
\IEEEauthorblockA{\textit{Fudan University} \\
}
\and
\IEEEauthorblockN{Xiangyang Xue\textsuperscript{1} }
\IEEEauthorblockA{\textit{Fudan University} \\
}
\and
\IEEEauthorblockN{Jianfeng Feng\textsuperscript{1} }
\IEEEauthorblockA{\textit{Fudan University} \\
}
}
\fi

\maketitle

\begin{abstract}
Monocular 3D object detection aims to predict the object location, dimension and orientation in 3D space alongside the object category given only a monocular image.
It poses a great challenge due to its ill-posed property which is critically lack of depth information in the 2D image plane.
While there exist approaches leveraging off-the-shelve depth estimation or relying on LiDAR sensors to mitigate this problem, the dependence on the additional depth model or expensive equipment severely limits their scalability to generic 3D perception. 
In this paper, we propose a stereo-guided monocular 3D object detection framework, dubbed SGM3D, adapting the robust 3D features learned from stereo inputs to enhance the feature for monocular detection.
We innovatively present a multi-granularity domain adaptation (MG-DA) mechanism to exploit the network's ability to generate stereo-mimicking features given only on monocular cues. 
Coarse BEV feature-level, as well as the fine anchor-level domain adaptation, are both leveraged for guidance in the monocular domain.
In addition, we introduce an IoU matching-based alignment (IoU-MA) method for object-level domain adaptation between the stereo and monocular predictions to alleviate the mismatches while adopting the MG-DA.
Extensive experiments demonstrate state-of-the-art results on KITTI and Lyft datasets.
Code and models will be made publicly available at \url{https://github.com/zhouzheyuan/sgm3d}.
\end{abstract}

% \begin{IEEEkeywords}
% Autonomous Vehicle Navigation,  Deep Learning for Visual Perception, Monocular 3D Object Detection
% \end{IEEEkeywords}

\section{Introduction}

\IEEEPARstart{3}{D} object detection is a fundamental and challenging task in computer vision as it allows to perceive the location, dimension, orientation in the 3D space alongside the category of objects, given only a single RGB input image.
It plays a critical role in numerous applications requiring some degrees of reasoning about object of interest, such as autonomous driving, visual navigation and robotics.

Despite the great success of 2D object detection~\cite{ren2015faster}, 3D object detection remains a largely unsolved problem due to its ill-posed property of critically lacking of depth information.
Lidar-based approaches are able to solve this dilemma as the inherent accuracy of the 3D structural and geometric knowledge are obtained by the LiDAR sensor.
However, the high cost of LiDAR limits its scalability.
As an alternative, there is a surge of interest in camera-based solution~\cite{brazil2019m3d, chen2020monopair, zhou2019objects,  vianney2019refinedmpl, ding2020learning, liu2021yolostereo3d, Ma2021delving, monorun2021, kumar2021groomed, Shi_2021_ICCV, wang2021depth, liu2021ground, wang2021progressive, CaDDN, Zou_2021_ICCV, Liu_2021_ICCV, Zhou_2021_CVPR, zhang2021objects, lu2021geometry}.
Camera-based methods can be categorized into stereo-based and monocular-based groups.
The former method leverages stereo image inputs with stereo matching and then either generate pseudo-LiDAR point clouds \cite{wang2019pseudo} that are feed into a LiDAR-based detector, or learn a implicit 3D representation for end-to-end localization of the 3D objects \cite{li2019stereo,guo2021liga}.
In contrast, monocular 3D object detection relies on a single image input that lacks of accurate depth information. 
Consequently, there is a clear performance gap between monocular-based methods and stereo-/LiDAR-based counterparts.
In summary, the key challenge of monocular 3D object detection lies in the robustness to inaccurate depth prediction.

\begin{figure*}
 \centering
 \includegraphics[width=\linewidth]{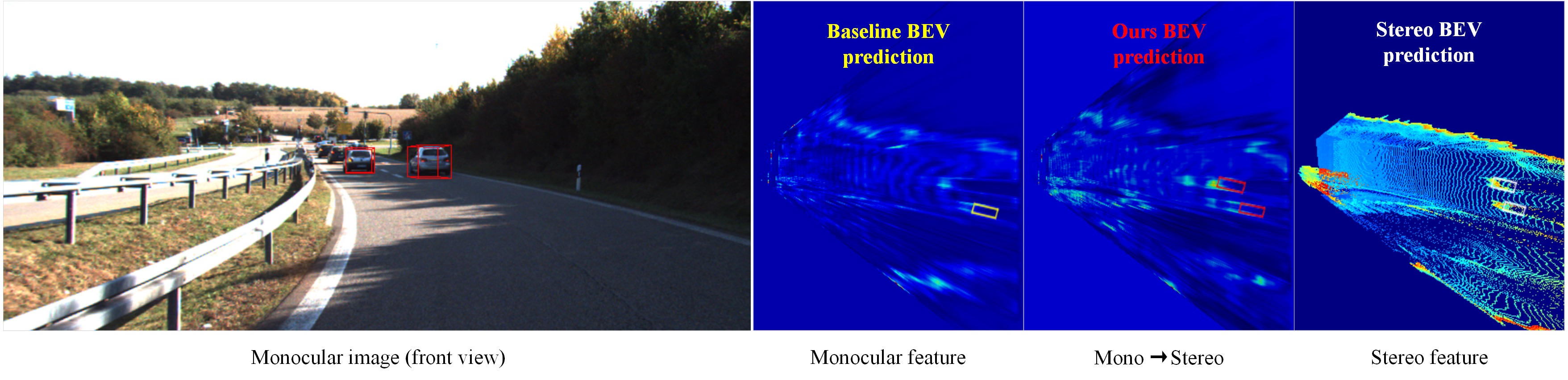}
 \caption{The qualitative comparison of the monocular feature (baseline), our stereo-guided monocular feature (``Mono $\rightarrow$ Stereo'' denotes adapting the feature from the monocular domain to the stereo domain), and the stereo feature. We average the channel of the feature after the BEV encoder in both the stereo and the monocular branch to obtain the response heatmap for visualization. 
 The response of the foreground regions marked by the red boxes is enhanced (i.e., getting warmer) with the guidance of the stereo feature marked by the white boxes compared with the vanilla monocular features marked by the yellow boxes.
 }
 \label{fig_intro}
\end{figure*}

In this paper, we propose to solve the monocular 3D detection by introducing a new framework with depth cues and without bringing any extra cost in the inference phase.
Previous works \cite{hinton2015distilling, Ye_2020_ECCV, xu2021spg} have proven that the feature adaptation approach can be used to exploit a network's ability to generate robust pseudo features based on the upstream fragile features.
In light of this statement, we propose to learn 3D detection features by transferring knowledge from the stereo counterpart.
Adapting from the Lidar-based pipeline is another alternative.
However,
(i) the inherent characteristic originating from camera differs substantially against Lidar.
(ii) expensive Lidar sensor prevent its potential piratical usage.
As shown in Figure~\ref{fig_intro}, with the guidance of stereo bird's-eye-view (BEV) feature, 
pseudo-stereo feature containing finer details of the objects can be generated, which is crucial for 3D detection.

However, naively transfer the stereo feature to the monocular branch may suffer from the imbalanced adaptation problem.
We therefore introduce a multi-granularity domain adaptation (MG-DA) mechanism to encourage consistency in different network stages to overcome the above-mentioned challenges.
The MG-DA enables our model to efficiently use the well-encoded geometric information from the strong stereo branch at both the coarse feature-level and the fine anchor-level to guide the learning process of the monocular branch.
Strictly anchor-level domain adaptation forces the anchors at the same location are matched (see Figure~\ref{fig_method}). 
However, a specific object might correspond to different anchors, which inevitably leads to inconsistent adaptation.
We present an IoU matching-based aligment (IoU-MA) method to align the predictions that across different anchors in both the stereo and monocular branches.
It improves detection performance not only by reducing the matching error of anchor-level domain adaptation but also by refining the box regression process.

To the best of our knowledge, we are the first to leverage stereo knowledge to guide a monocular 3D object detection network, which fundamentally enhances the monocular performance without introducing any extra cost during inference. Our main contributions are summarized as follows:
\begin{itemize}
\item We propose a stereo-guided monocular 3D object detection framework named SGM3D, which forces the network to actively mimic stereo representations based only on monocular images.
\item We introduce a multi-granularity domain adaptation (MG-DA) mechanism that guides the monocular detector in hierarchical stages, including at the feature-level and the anchor-level.
\item We present an IoU matching-based alignment (IoU-MA) method for object-level domain adaptation between the two modalities to further optimize the learning process.
\item The proposed SGM3D framework achieves state-of-the-art performance on the challenging KITTI \cite{geiger2012we} and Lyft \cite{lyft} datasets.
\end{itemize}
\section{Related work}
\subsection{Image-only 3D detection} Standard monocular 3D detection approaches solely rely on RGB images to predict 3D bounding boxes. 
To solve this ill-posed problem caused by the inherent scale and depth ambiguity, most existing methods utilize the 2D-3D geometric constraints to improve the representation ability of the models. 
M3D-RPN \cite{brazil2019m3d} proposes a standalone 3D region proposal network to leverage prior statistics, which serves as a strong initial guess for each 3D bounding box. 
Monopair \cite{chen2020monopair} enhances the modeling capability on occlusive objects by encoding spatial constraints between partially-occluded objects and their adjacent neighbors. 
Inspired by CenterNet \cite{zhou2019objects}, recent progresses \cite{zhang2021objects, lu2021geometry} formulate the image-only 3D detection as an anchor-free detection problem, and further combine the projection constraint to assist in 3D box construction. 
These methods predict the center location, dimension, and the rotation of the 3D box in the image plane, and then solve the corresponding 3D properties with nonlinear least-squares optimization.

\subsection{Depth-assisted 3D detection} Considering the difficulty in perceiving 3D properties from 2D images, many methods take pixel-wise depth maps generated by off-the-shelf monocular depth estimators as an additional input. Some pioneering works \cite{vianney2019refinedmpl} adopt a pseudo-LiDAR framework to transform the monocular image into 3D space with estimated depth, and then utilize LiDAR-based detectors to estimate 3D Boxes. Despite the promising performance, these approaches rely heavily on the accuracy of the estimated depth and may not be well generalizable to new scenarios. On the other hand, $\rm D^4LCN$ \cite{ding2020learning} proposes to utilize depth maps as a guidance to learn the local kernels from RGB images by dynamic-depthwise-dilated local convolutions. DDMP-3D \cite{wang2021depth} further presents a graph-based formulation to effectively integrate the multi-scale depth information and learn depth-aware feature representations by adaptively sampling context-aware nodes within the RGB context.

\subsection{Domain adaptation} Domain adaptation (DA), was originally proposed to address the domain shift problem, has drawn great attention for many computer vision tasks \cite{zhao2019geometry,pinheiro2019domain}. 
These methods aim to mitigate the domain gap and enhance the generality ability of the neural networks. 
In addition, it is found that some tasks are easy to solve in one domain while being difficult in another.
Since the LiDAR sensor and the stereo camera can provide much more accurate geometry information than a single camera, performing domain adaptation between the monocular and stereo (LiDAR) modalities is a naive and effective way to boost the performance.
In this paper, we perform feature adaptation from the monocular domain to the stereo domain on hierarchical levels, which enforces the monocular 3D object detector to generate stereo-mimic features that are much more favorable and robust for the 3D object detection task. 
\section{Methodology}
\begin{figure*}
 \centering
 \includegraphics[width=\linewidth]{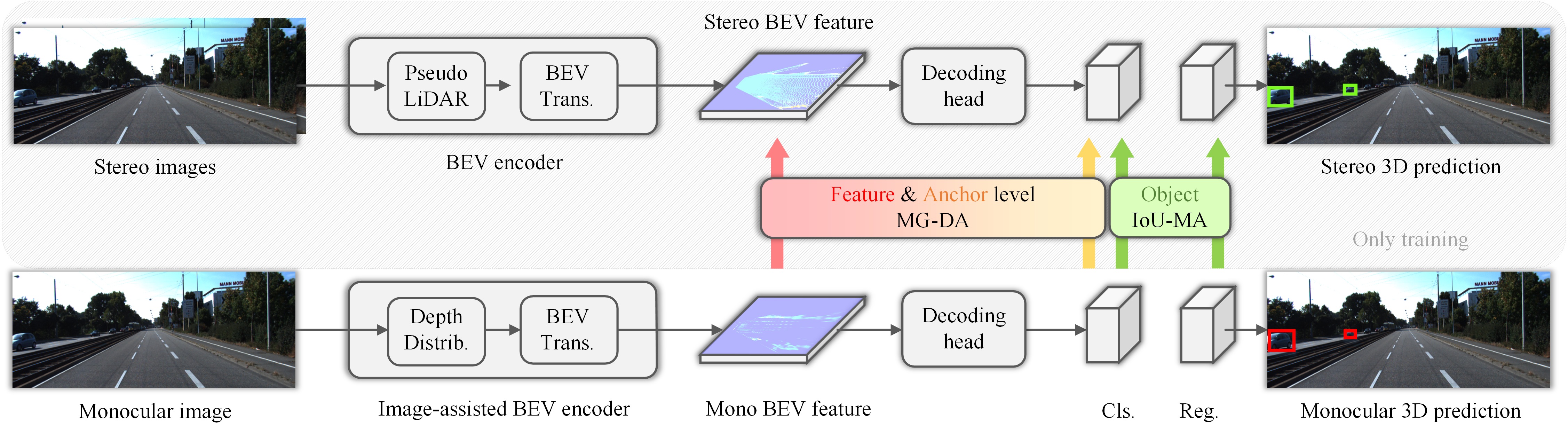}
 \caption{The overview of the proposed SGM3D. The upper branch takes stereo images as the input and obtains the BEV features via converting the stereo images to pseudo-LiDAR point clouds. The bottom branch takes the monocular image as the input and obtains the BEV features via the image-assisted BEV encoder, which will be explained in Sec.~\ref{i2b}. We adapt the features of different levels from the monocular domain to the stereo domain through our MG-DA, including the coarse intermediate feature-level, fine-grained anchor-level on the classification head. We further use IoU-MA to compensate for the mismatch problem lying in the previous anchor-level domain adaptation process and perform object-level adaptation. The predicted 3D boxes of the stereo branch and our method are drawn in green and red, respectively.
 }
 \label{fig_main}
\end{figure*}
Due to the additional geometric constraints and extra contextual knowledge, stereo 3D object detection usually yields higher performance than monocular approaches.
To improve the performance of monocular 3D object detection, in this letter, we take advantage of stereo representation and propose a simple yet effective domain-adaptation-based framework named SGM3D.
First, an overview of the framework is introduced in Sec.~\ref{overview}.
Next, for feature adaptation at both the coarse feature-level and the fine anchor-level, we propose our MG-DA, which is explained in detail in Sec.~\ref{mg}.
Then, we note that due to the inconsistent depth estimation and different features in the stereo and monocular domains, predicted locations with respect to the same anchor may differ from each other. 
To alleviate this mismatch problem at the anchor-level, we present our IoU-MA, which is introduced in Sec.~\ref{iou}.
Finally, in Sec.~\ref{loss}, the objective loss functions of SGM3D are presented.

\subsection{Framework overview}
\label{overview}
We propose a stereo-guided monocular 3D object detection (SGM3D) approach that leverages the representations learned from stereo images to guide monocular-based 3D object detection learning in a multi-granular manner.
An overview of the framework is illustrated in Figure~\ref{fig_main}.
The upper part represents the stereo 3D object detection branch, which is used only during training, and the bottom part is the monocular 3D object detection branch.
During the training process, our network takes both stereo image pairs and monocular images as inputs to the two branches.

\noindent \textbf{In the stereo branch}, the corresponding depth is learned by a pretrained stereo matching model, PSMNet \cite{chang2018pyramid}, and we convert the pixel points into 3D pseudo point clouds based on the estimated depth and camera intrinsics.
Then, we utilize the lightweight PointPillars \cite{Lang_2019} feature encoder to generate BEV feature maps.

\noindent \textbf{In the monocular branch}, we first feed a red--green--blue (RGB) input image into an encoder to learn high-level feature representations and leverage a dense depth distribution prediction module to map the image features into 3D space.
Frustum features are generated through pixelwise multiplication of the dense predicted depth distribution and the image features.
Using the camera intrinsic parameters, the frustum features are lifted into 3D space. Then, the 3D spatial features are collapsed into BEV features for further 3D object regression.
\label{i2b}

Note that the two branches have similar network architectures, except for the transformation of 2D images into BEV features.
During inference, given a single image as input, we retain only the monocular branch for prediction, which incurs no additional computational cost.

\subsection{Multi-granularity domain adaptation (MG-DA) mechanism }
\label{mg}
The MG-DA encourages consistency between the outputs from the stereo and monocular branches in the intermediate feature representations and the predictions per anchor.

\subsubsection{Coarse feature-level domain adaptation}
\label{feature}
Since the depth information in the stereo branch is typically more accurate than that in the monocular branch, the corresponding 3D scene representations are generally more informative. Therefore, it is beneficial to encourage the image-to-BEV encoder in the monocular branch to learn from its stereo counterpart.

We first align the features at a coarse level.
As mentioned above, the BEV features in the two branches are spatially aligned, but due to differences in the estimated depths, these features may vary greatly.
Therefore, we exploit an L2-distance-based loss to minimize the BEV feature distances between the stereo and monocular branches.
To ensure stability of the domain adaptation at the intermediate feature-level, we focus only on foreground regions where a foreground object exists.
The parameters of the monocular branch are tuned to encourage the generation of features that more closely resemble the corresponding stereo features to promote robustness to noise in the depth estimation.
The foreground mask is computed by projecting the ground-truth 3D bounding boxes onto the BEV view and resizing the boxes to match the shapes in the BEV feature map.
The coarse feature-level domain adaptation loss is expressed as follows:
\begin{equation}
  \mathcal{L}_{feature}= (||\mathcal{F}_{}^{M} - \mathcal{F}_{}^{S}||^2) \cdot \mathcal{M}_{fg},
 \label{loss_feature}
\end{equation}
where $\mathcal{F}_{}^{M}$ and $\mathcal{F}_{}^{S}$ are the two feature maps generated by the BEV encoder of the monocular branch and the stereo branch, $\mathcal{M}_{fg}$ is the foreground feature mask.

\subsubsection{Fine anchor-level domain adaptation}
\label{anchor}
In addition to intermediate feature alignment, the anchor-level consistency between the stereo and monocular branches is further explored at the outputs of each decoder.
Instead of directly adopting the hard ground-truth labels for supervision, many pioneering works \cite{wang2019distilling, guo2021distilling, zheng2021se} have proven that labels generated from a better domain can provide additional knowledge guidance and help to reveal the differences between samples at a more fine-grained level.
To clearly describe the domain adaptation process, we use $A_{ijk}$ to represent the $k$-th anchor located at ($i, j$) in the image coordinates and $P_{ijk}$ to represent the prediction at $A_{ijk}$.

Given that the stereo and monocular branches share the same anchor-based method, $A_{ijk}^{S}$ in the stereo branch is aligned with $A_{ijk}^{M}$ in the monocular branch. We perform anchorwise alignment of the classification predictions as follows.
First, we extract the foreground anchors by matching them with ground-truth bounding boxes, and the remaining anchors are designated as background anchors.
Then, the Kullback-Leibler(KL) divergence loss is adopted to regularize the distances between the predicted classification scores of each anchor.
Specifically, the raw predicted classification scores are normalized to the probability distribution using the softmax function:
\begin{equation}
\begin{aligned}
\mathcal{L}_{anchor}^{} = {\lambda_{fg}^{}} (\frac{1}{|\mathcal{A}_{fg}|}\sum_{i \in A_{fg}}^{} c_{i}^S \log_{}{\frac{c_{i}^S}{c_{i}^M}}) \\
+  {\lambda_{bg}^{}} (\frac{1}{|\mathcal{A}_{fg}|}\sum_{i \in A_{bg}}^{} c_{i}^S \log_{}{\frac{c_{i}^S}{c_{i}^M} }),
\end{aligned}
\label{loss_anchor}
\end{equation}
where $c_{i}^S$ and $c_{i}^M$ are the $i$-th classification scores after applying the softmax function to the raw classification head for stereo and monocular branches, respectively.
$\mathcal{A}_{fg}$ and $\mathcal{A}_{bg}$ denote the set of foreground anchors and background anchors.
$\lambda_{fg}^{}$ and $\lambda_{fg}^{}$ represent different loss weights for foreground and background proposal anchors for loss balancing.

\subsubsection{Loss function}
By combining the coarse feature-level domain adaptation loss and the fine anchor-level domain adaptation loss, the overall loss function of the MG-DA is formulated as follows:
\begin{equation}
    \mathcal{L}_{MG-DA} = 
    \lambda_{feature}\mathcal{L}_{feature} +
    \lambda_{anchor}\mathcal{L}_{anchor},
\label{loss_mg}
\end{equation}
where $\lambda_{feature}$ and $\lambda_{anchor}$ represent the weights of the different losses.

\subsection{IoU matching-based alignment (IoU-MA) method}
\label{iou}
Through the abovementioned coarse feature-level and fine anchor-level alignment processes, the network is encouraged to learn more reliable and stereo-mimicking feature representations and to produce similar classification confidences for each class.
However, in anchor-level alignment, domain adaptation is performed strictly with the same anchor, which introduces additional error when two predicted locations have a large offset.

\begin{figure}
 \centering
 \includegraphics[width=\linewidth]{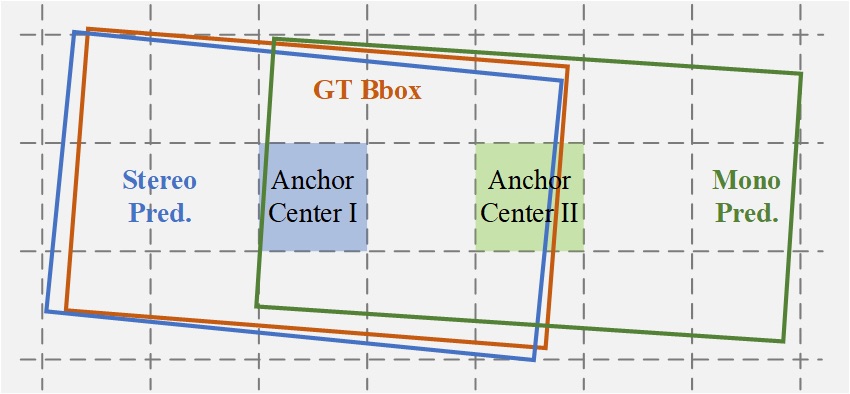}
 \caption{
The illustration of the mismatch issue in the anchor-level domain adaptation. 
For a specific object, the predictions from the stereo and the monocular branches may not originate from the same anchor. 
The stereo prediction box colored in blue is originated from ``Anchor center I'', while the monocular prediction colored in green is originated from ``Anchor center II''.
Therefore, it leads to an inconsistent alignment for adaptation, which influences the performance.
}
\label{fig_method}
\end{figure}

Specifically, as shown in Figure~\ref{fig_method}, we denote the stereo prediction (blue) by $P_{ijk}^{S}$ and the monocular prediction (green) by $P_{i^{'}j^{'}k{'}}^{M}$.
$P_{ijk}^{S}$ is a foreground object, while the monocular prediction $P_{ijk}^{M}$ at this anchor is ignored due to its low score.
On the other hand, $P_{i^{'}j^{'}k^{'}}^{M}$ is another foreground prediction from the monocular branch that overlaps with $P_{ijk}^{S}$.
Accordingly, due to discrepancies caused by different depth estimates, the same target object may be associated with different anchors in different locations.

To alleviate this problem and encourage further fine-grained domain adaptation, we specifically design our IoU-MA.
To obtain high-confidence predictions, we first filter the predictions by the confidence scores from the classification head. 
Then, for each predicted 3D box in the monocular branch, the best matching stereo prediction box (with the highest IoU score) is selected. 
Furthermore, the stereo predictions can introduce disturbances, with the potential to improve the robustness of the network.

For the generated paired bounding boxes, we adopt a smooth L1-distance-based loss to minimize the difference in the predicted confidences of the classification scores in the stereo and monocular domains ($L_{object}^{cls}$):
\begin{equation}
\mathcal{L}_{object}^{cls} = \frac{1}{|\mathcal{A}_{match}|} \sum_{i \in \mathcal{A}_{match}}^{} {\mathcal{L}_{smooth-L1}}(c_{i}^{M}, c_{i}^{S}),
\label{loss_oc}
\end{equation}
where ${\mathcal{A}_{match}}$ represents the set of matched objects in prediction, $c_{i}^{M}$ and $c_{i}^{S}$ are the $i$-th classification prediction scores for monocular and stereo branch after applying truncation functions to them.

Next, we apply the L2 distance-based loss for paired bounding boxes from the stereo and monocular branches ($L_{object}^{box}$) to minimize the misalignment errors, so as to alleviate the geometric mismatch:
\begin{equation}
\mathcal{L}_{object}^{box} =  \frac{1}{|\mathcal{A}_{match}|} \sum_{i \in \mathcal{A}_{match} }^{} \mathcal{L}_{L2}(res_{i}^{M},res_{i}^{S} ) ,
\label{loss_ob}
\end{equation}
where ${\mathcal{A}_{match}}$ represents the set of matched objects in prediction, $res_{i}^{M}$ and $res_{i}^{S}$ denote the residual object parameters of monocular branch and stereo branch, including the center position, size, and orientation of a 3D bounding box.

Combining the losses for classification score and box regression together, we generate IoU-MA loss as:
\begin{equation}
\mathcal{L}_{IoU-MA} = \lambda_{object} (\mathcal{L}_{object}^{cls} + \mathcal{L}_{object}^{box}),
\label{loss_iou}
\end{equation}
where $\lambda_{object}$ represent the loss weights for IoU-MA.

\subsection{Objective functions}
\label{loss}
Combining the MG-DA loss (Equation~\ref{loss_mg}) and the IoU-MA loss (Equation~\ref{loss_iou}) together, the overall stereo guided monocular domain adaptation loss is formulated as follows:
\begin{equation}
    \mathcal{L}_{SGM} = 
    \mathcal{L}_{MG-DA} +
    \mathcal{L}_{IoU-MA} .
\label{loss_da}
\end{equation}

In addition to the domain adaptation losses, there are three losses for the standard object detection task, including the focal loss $\mathcal{L}_{cls}$ for classification , the smooth-L1 loss $\mathcal{L}_{box}$ for 3D box regression and the binary cross-entropy loss $\mathcal{L}_{dir}$ for direction classification following SECOND \cite{yan2018second}.

Inspired by LSS\cite{philion2020lift} and CaDDN\cite{CaDDN}, in the image-assisted BEV encoder of the monocular branch (Sec.~\ref{i2b}), we divide the predicted depth range into bins to convert the depth regression into a classification problem. 
Therefore, we train a depth distribution module by classifying each pixel in the image features into the correct bins, using the focal loss for balancing the foreground and background pixels:
\begin{equation}
\label{loss_depth}
    \mathcal{L}_{depth} =  \frac{1}{|\mathcal{F}|}\sum_{p \in \mathcal{F}}^{} -\alpha(1-d_{p})^\gamma log(d_{p}),
\end{equation}
where $\mathcal{F}$ denotes the set for the image features, $d_{p}$ represents the probability for depth bins for a pixel in the image features, $\alpha$ and $\gamma$ are the parameters of the focal loss.

We denote the monocular branch model loss of the 3D object detection task as $\mathcal{L}_{3Ddet}$:
\begin{equation}
\label{loss_base}
\begin{aligned}
  \mathcal{L}_{3Ddet}= \frac{1}{N_{pos}}(\lambda_{cls} \mathcal{L}_{cls} + \lambda_{box} \mathcal{L}_{box} + \lambda_{dir} \mathcal{L}_{dir})\\ +\lambda_{depth} \mathcal{L}_{depth},
\end{aligned}
\end{equation}
where $N_{pos}$ is the number of positive anchors, $\lambda_{cls}$, $\lambda_{box}$,  $\lambda_{dir}$, $\lambda_{depth}$ represent the loss weights for different sub task.

With the stereo guided monocular domain adaptation loss (Equation~\ref{loss_da}) and the 3D object detection loss (Equation~\ref{loss_base}) mentioned above, the total loss of our network is formulated as:
\begin{equation}
    \mathcal{L}_{} =  \mathcal{L}_{SGM} + \mathcal{L}_{3Ddet},
\end{equation}
where we set the same weight for both terms.

\section{Experiments}
\subsection{Implementation}
\subsubsection{Datasets}
To fully validate the effectiveness of our proposed SGM3D, we conduct extensive experiments on the two most challenging datasets: KITTI and Lyft datasets.
The KITTI dataset contains 7,481 and 7,518 images for training for testing, respectively.
Following SECOND, we utilize the train-val split of the KITTI dataset to evaluate the proposed SGM3D. 
We also conduct experiments on the Lyft dataset, which includes 22,680 frontal-view images. 
The dataset is randomly separated into 18144 images for training and 4536 images for validation, including three categories (``Car'', ``Pedestrian'', and ``Cyclist''). 
\subsubsection{Evaluation metrics}
Precision-recall curves are utilized for evaluation, and we report the average precision (AP) results of 3D and Bird’s eye view (BEV) object detection of the KITTI and Lyft datasets.
Three levels of difficulty are defined in the benchmark according to the 2D bounding box height, occlusion, and truncation degree, namely, ``Easy'', ``Mod.'' and ``Hard''. 
\subsubsection{Training details}
We utilize 8 Nvidia Tesla v100 GPUs to train the model for 80 epochs on KITTI and 30 epochs on Lyft, respectively.
The learning rate is set to 0.001 with the cosine annealing learning rate strategy, and 0.01 is adopted for the learning rate decay. 
We use a batch size of 2, and the network is optimized by Adam with a momentum of 0.9.
We set $\lambda_{feature}$ = 0.1 and $\lambda_{anchor}$ = 1.0 in Equation~\ref{loss_mg}. In Equation~\ref{loss_iou}, we set $\lambda_{object}$ = 0.01. For the focal loss we set $\alpha$ = 0.25 and $\gamma$ = 2 following the original paper\cite{lin2017focal} settings. 
For the monocular branch model loss $\mathcal{L}_{base}$ in Equation~\ref{loss_base}, we use the same setting in SECOND of $\lambda_{cls}$ = 1.0, $\lambda_{box}$ = 2.0, $\lambda_{dir}$ = 0.2 and $\lambda_{depth}$ = 3.0, respectively.

\subsection{Comparison with state-of-the-art methods}
\subsubsection{Results on ``Car''}
Table \ref{tab_test} reports the results on the ``Car'' category on the KITTI test set at IoU = 0.7.
\begin{table*}[htb]
 \caption{Comparison with state-of-the-art methods on the KITTI \textbf{test} set for $\rm AP_{3D}$ and $\rm AP_{BEV}$ of "Car" at IoU = 0.7. Our SGM3D achieves new state-of-the-art performance in both accuracy and speed. ``-'' represents the unknown information.}
 \small
 \centering
 {
 \begin{tabular}{|C{3.2cm}|C{1.8cm}|c|C{1cm}C{1cm}C{1cm}|C{1cm}C{1cm}C{1cm}|c|}
  \hline
  \multirow{2} * {Method} & \multirow{2} * {Reference} & \multirow{2} * {FPS}& \multicolumn{3}{c|}{$\rm AP_{3D}$} & \multicolumn{3}{c|} {$\rm AP_{BEV}$}      &  \multirow{2} * {GPU}                \\
    &           & & Mod.          & Easy          & Hard          & Mod.          & Easy          & Hard        &  \\
    \hline
    \hline
    YOLOMono3D\cite{liu2021yolostereo3d} &ICRA2021 &13 &12.06 & 18.28 & 8.42 &17.15 &27.94 &16.00 &GTX 1080Ti \\
    Monodle\cite{Ma2021delving} &CVPR2021 &25 &12.26 &17.23 &10.29 &18.89 &24.79 &16.00 &GTX 1080Ti\\
    MonoRUn\cite{monorun2021} &CVPR2021 &14 &12.30 &19.65 &10.58 &17.34 &27.94 & 15.24 &RTX 2080Ti \\
    GrooMeD-NMS\cite{kumar2021groomed} &CVPR2021 &8 & 12.32 &18.10 &9.65 &18.27 & 26.19 & 14.05 &Titan X \\
    MonoRCNN\cite{Shi_2021_ICCV} &ICCV2021 &14 & 12.65 & 18.36 & 10.03 & 18.11 & 25.48 & 14.10 & Titan X\\
    DDMP-3D\cite{wang2021depth}    &CVPR2021       & 6         & 12.78 & 19.71 & 9.80 & 17.89 & 28.08 & 13.44 & Tesla V100 \\
    GAM3D\cite{liu2021ground} &RAL2021 &20 &13.25 &21.65 &9.91 &17.98 &29.81 &13.08 &GTX 1080Ti \\
    PCT\cite{wang2021progressive}        &NIPS2021       & 22         & 13.37 & 21.00 & 11.31 & 19.03 & 29.65 & 15.92 & Tesla V100 \\
    CaDDN\cite{CaDDN}      &CVPR2021       & 33         & 13.41 & 19.17  & 11.46 & 18.91 & 27.94 & 17.19 &Tesla V100 \\
    DFR-Net\cite{Zou_2021_ICCV}    &ICCV2021       & 6 & 13.63 & 19.40 & 10.35 & 19.17 & 28.17 & 14.84 & GTX 1080Ti \\ 
    AutoShape\cite{Liu_2021_ICCV} &ICCV2021 &25 &13.72 & 21.75 & 10.96 & 19.00 & 30.43 & 15.57 & Tesla V100 \\
    MonoEF\cite{Zhou_2021_CVPR} &CVPR2021 &33 & 13.87 & 21.29 & 11.71 & 19.70 & 29.03 & 17.26 &- \\ 
    MonoFlex\cite{zhang2021objects}    &CVPR2021       & 29          & 13.89 & 19.94  & 12.07 &19.75 &28.23  &16.89 &RTX 2080Ti \\
    GUP Net\cite{lu2021geometry} &ICCV2021 &29 &14.20 &20.11 &11.77 &- &- &- & TiTan XP \\
    \hline
    Ours        &-              & 33         & \textbf{14.65} & \textbf{22.46} & \textbf{12.97} & \textbf{21.37} &\textbf{31.49}&\textbf{18.43} & Tesla V100 \\
  \hline
 \end{tabular}
 }
 \label{tab_test}
\end{table*}

On the KITTI leaderboard, the proposed SGM3D framework ranks at the top among all monocular-based 3D object detection methods.
Compared with the cutting-edge Monoflex framework, our method achieves superior results on all settings (``Easy'', ``Mod.'' and ``Hard'') while maintaining a real-time speed of 33 FPS on a Tesla V100 GPU, being 5 times faster than DDMP-3D.
Note that some methods \cite{ding2020learning, wang2021depth} use a pseudo depth map as additional input for 3D object detection, while our end-to-end method does not require any extra off-the-shelf depth estimation network and outperforms those methods by a large margin.

\subsubsection{Results on ``Pedestrian'' and ``Cyclist''}
Different from the ``Car'' category, ``Pedestrian'' and ``Cyclist'' have nonrigid structures and relatively small scales; therefore, they are much more challenging targets for 3D object detection.
\begin{table}[htb]
\caption{3D object detection ($\rm AP_{3D}$) performance for $\rm AP_{3D}$ of ``Pedestrian'' and ``Cyclist'' on the KITTI \textbf{test} set.}
\small
\centering
\resizebox{\linewidth}{!}
 {
\begin{tabular}{|c|ccc|ccc|}
\hline
    \multirow{2} * {Method} & \multicolumn{3}{c|}{Pedestrian} & \multicolumn{3}{c|} {Cyclist}   \\
    & Mod.  & Easy  & Hard  & Mod.  & Easy  & Hard  \\
\hline
\hline
    $\rm D^4LCN$\cite{ding2020learning}     & 3.42  &4.55& 2.83                             & 1.67 & 2.45 & 1.36\\ 
    DDMP-3D\cite{wang2021depth}                   & 3.55 & 4.93  & 3.01                           & 2.50	& 4.18  & 2.32\\
    MonoFlex\cite{zhang2021objects}                    & 6.31  & 9.43  & 5.26                          & 2.35  & 4.17	& 2.04\\

    \hline
    Ours                     & \textbf{8.81} &\textbf{13.99} & \textbf{7.26} & \textbf{2.92} & \textbf{5.49} & \textbf{2.64}\\
\hline
\end{tabular}
}
\label{tab_pc}
\end{table}

Table~\ref{tab_pc} reports the $\rm AP_{3D}$ values for ``Pedestrian'' and ``Cyclist'' on the test set at IoU = 0.5.
Our SGM3D framework achieves new state-of-the-art performance on both the ``Pedestrian'' and ``Cyclist'' categories.

\subsubsection{More qualitative results}
\begin{figure}[htb]
 \begin{center}
  \includegraphics[width=\linewidth]{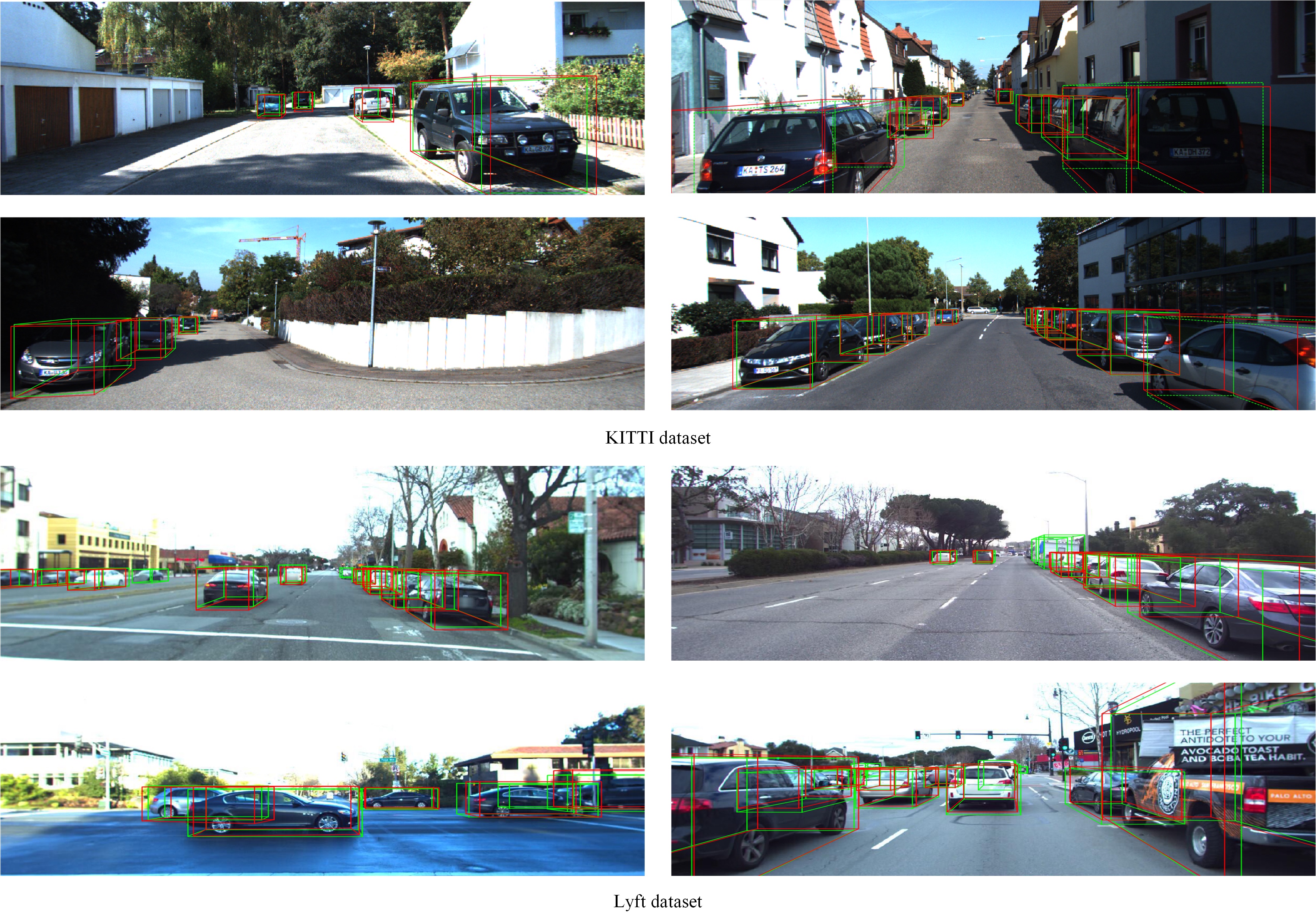}
 \end{center}
 \caption{Visualization of the results on the KITTI and the Lyft validation split sets. The ground-truth 3D boxes and the predicted 3D boxes of our method are drawn in green and red, respectively.}
 \label{fig_result}
\end{figure}

\begin{figure}[htb]
 \begin{center}
  \includegraphics[width=\linewidth]{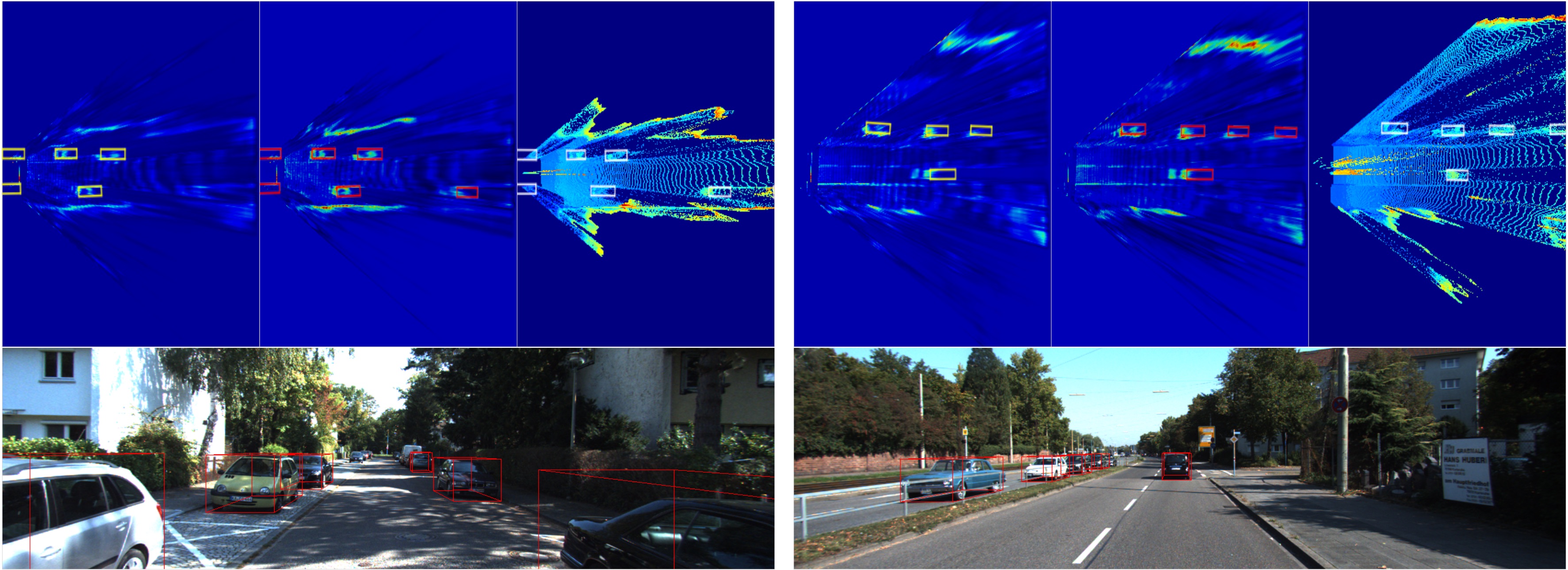}
 \end{center}
 \caption{Visualization to the BEV feature maps and result on the KITTI validation split set. The predicted 3D boxes of baseline and our method, ground-truth 3D boxes are drawn in yellow, red and white, respectively. The three columns are the monocular feature (baseline), our stereo-guided monocular feature, and the stereo feature, respectively.}
 \label{fig_result_bev}
\end{figure}

Figure~\ref{fig_result} presents more qualitative results on the KITTI and Lyft validation datasets.
The ground truth and the results of our method are colored green and red, respectively.
As shown in Figure~\ref{fig_result}, the network is able to detect challenging occluded and distant objects due to the proposed MG-DA and IoU-MA.
To further observe how domain adaptation fundamentally improves 3D object detection, we visualize the BEV feature maps obtained from the monocular features (baseline), our stereo-guided monocular features, and the stereo features as well as their corresponding detection results in Figure~\ref{fig_result_bev}. We average each feature map along the channel dimension to obtain the corresponding response heatmap, in which warmer colors indicate higher response. The pixelwise differences and differences in prediction ability between the stereo and monocular features are large due to the inaccurate depth estimation in the monocular branch. As depicted in the first and second columns for each scene, distant and occluded objects are much more obvious and can be easily detected after our model performs effective adaptation from the monocular domain to the stereo domain. This visualization demonstrates that our model can generate robust features for 3D object detection.


\subsection{Ablation study}
\subsubsection{Main ablative analysis of different domain adaptation levels}
The ablation experiments in Table \ref{tab_ablation} demonstrate the effectiveness of different settings:
(a) the baseline (adopting only the monocular 3D detection branch);
(b) feature-level domain adaptation (coarse domain adaptation from monocular BEV features to stereo BEV features);
(c) anchor-level domain adaptation (fine domain adaptation based on anchors);
(d) object-level domain adaptation (fine domain adaptation based on object matches between the two branches);
(e) the MG-DA with both feature-level and anchor-level domain adaptation; and
(f) our full approach.

\begin{table}[htb]
 \caption{Ablative analysis on the KITTI validation set for $\rm AP_{3D}$ of "Car".
 }
 \small
 \centering
 \resizebox{\linewidth}{!}
{
 \begin{tabular}{|c|c|c|c|ccc|}
  \hline
  \multirow{3} * {Group} & Feature-level & Anchor-level & Object-level & \multicolumn{3}{c|}{$\rm AP_{3D}$}\\
  & (Sec.~\ref{feature})  & (Sec.~\ref{anchor})  & (Sec.~\ref{iou}) & Mod. & Easy & Hard\\
  \hline
  \hline
    (a)       & -             & -             & -             & 14.50 & 20.98 & 12.27 \\
    (b)      & \checkmark    & -             & -             & 15.78 & 22.85 & 13.31 \\
    (c)    & -             & \checkmark    & -              &16.65 &24.79 &13.92 \\
    (d)     & -             & -             & \checkmark    & 16.05 & 23.01 & 13.36 \\
    (e)       & \checkmark    & \checkmark    & -             & 16.94 & 25.53 & 14.75 \\
    (f)    & \checkmark    & \checkmark    & \checkmark    &\textbf{17.81} & \textbf{25.96} & \textbf{15.11} \\
  \hline
 \end{tabular}
 }
 \label{tab_ablation}
\end{table}

From Table \ref{tab_ablation}, we can observe the following phenomena.
First, as seen by comparing (b), (c), and (d) with (a), the feature-level, anchor-level, and object-level domain adaptation processes result in improvements over the baseline by 1.28, 2.15, and 1.55, respectively, in the moderate setting, thus confirming that performing adaptation at each of these levels individually is necessary for the 3D object detection task.
Next, as seen by comparing (e) with (a), a noticeable gain of 2.44 is achieved with our MG-DA (``feature-level'' + ``anchor-level'') in the moderate setting, which shows that adaptation at both of these levels is critical to the model.
Finally, as seen by comparing (f) with (e), further applying the IoU-MA to the network leads to an additional performance gain of 0.87 in the moderate setting, thus confirming its effectiveness in increasing detection accuracy by aligning the same targets originating from different anchors or locations.
In summary, Table \ref{tab_ablation} clearly demonstrates the effectiveness of the proposed MG-DA and IoU-MA.
\subsubsection{Domain adaptation between different modalities}
\label{lidar}
To further validate the effectiveness and generalization ability of our SGM3D framework, we performed domain adaptation between different data modalities on the KITTI validation set.
In Table \ref{tab_modality}, the first and second rows report the results of using LiDAR and pseudo-LiDAR data (generated from stereo images) separately as the input to the PointPillars.
The third row shows the baseline results. 
The fourth and fifth rows show the results after performing domain adaptation between different modalities.
\begin{table}[htb]
 \caption{Results on $\rm AP_{3D}$ and $\rm AP_{BEV}$ via domain adaptation between different data modalities of "Car" on KITTI val set.
 ``LiDAR'', ``Stereo'' and ``Mono'' represent using the single modality data for training without domain adaptation. ``$\rightarrow$'' denotes the adaptation direction between different modalities.}
 \small
 \centering
 \resizebox{\linewidth}{!}
 {
 \begin{tabular}{|c|ccc|ccc|}
  \hline
  \multirow{2} * {Modality} & \multicolumn{3}{c|}{$\rm AP_{3D}$} & \multicolumn{3}{c|} {$\rm AP_{BEV}$}     \\
    & Mod.  & Easy  & Hard  & Mod.  & Easy  & Hard  \\
  \hline
  \hline
  LiDAR     & 76.04 & 87.19 & 74.46 & 87.11 & 91.44 & 84.96 \\
  Stereo    & 58.20 & 77.79 & 55.60 & 89.40 & 70.65 & 67.94 \\
  Mono      & 14.50 & 20.98 & 12.27 & 20.20 & 28.71 & 18.28 \\
  \hline
    Mono $\rightarrow$ LiDAR    & 16.41 & 23.55 & 13.85 & 23.16 & 32.99 & 20.11  \\
  Mono $\rightarrow$ Stereo & \textbf{17.81}  & \textbf{25.96} & \textbf{15.11} & \textbf{23.62} & \textbf{34.10} & \textbf{20.49} \\
  \hline
 \end{tabular}
 }
 \label{tab_modality}

\end{table}

As shown in Table~\ref{tab_modality}, the ``Mono $\rightarrow$ Stereo'' result is better than the ``Mono $\rightarrow$ LiDAR'' result. Accordingly, we believe that the domain gap between stereo and monocular features are much smaller than the gap between LiDAR and monocular features.
The purely structural features extracted from LiDAR point clouds may be quite different from the features generated based on depth cues from images (monocular or stereo), leading to greater difficulty in adaptation for the network.
Note that this experiment also demonstrates that the MG-DA can be used for both stereo and LiDAR settings.
In the future, we will also investigate different feature representations that may be more suitable for LiDAR-guided monocular 3D object detection, potentially outperforming the current stereo-guided approach.
\subsubsection{Different loss weights for foreground and background anchors}
\label{weights}
Since there is a large difference between the numbers of foreground and background anchors, to balance the anchor-based domain adaptation process, we adopt different foreground and background loss weights, as shown in Equation~\ref{loss_anchor}.
\begin{table}[htb]
 \caption{Comparison of different loss weights for foreground and background anchors on KITTI val set for $\rm AP_{3D}$ of "Car".
 }
 \small
 \centering
 {
 \begin{tabular}{|C{0.75cm}|C{0.75cm}|C{1.4cm}C{1.4cm}C{1.4cm}|}
  \hline
  \multirow{2} * {$\lambda_{fg}^{}$} & \multirow{2} * {$\lambda_{bg}^{}$} & \multicolumn{3}{c|}{$\rm AP_{3D}$} \\
     &  & Mod. & Easy & Hard\\
\hline
\hline
    - &-        &14.50 &20.98 &12.27 \\
    1 &1        &15.67  &24.40  &13.19 \\
    1 &0.5        &15.87  &24.19  &13.38 \\
    1 &0.25      &16.00  &23.27  &13.38       \\
    1 &0.05      &\textbf{16.65} &\textbf{24.79} &\textbf{13.92} \\
    1 &0.025     &15.71  &23.31  &13.30 \\
    
  \hline
 \end{tabular}
 }
 \label{tab_hyper_anchor}
\end{table}

The first row in Table \ref{tab_hyper_anchor} shows the results without adaptation.
Since foreground features contain much more crucial information for the adaptation process than background features do, we reduced the background loss ratio in a stepwise manner to search for the best value of this hyperparameter and found that a ratio of ``1:0.05'' yields the best results.
From the last two rows, we can see that too small a background loss ratio (1:0.025) leads to a performance drop, indicating that background regions also contain some important information for 3D object detection.
\subsection{Results on the Lyft dataset}
To further validate the generalization ability of our method, we conducted experiments on the Lyft dataset. Instead of using ResNet, as on the KITTI dataset, we used MobileNetV3-Large \cite{howard2019searching} as the backbone network due to computational resource limitations. Since the Lyft dataset does not contain stereo images, we used LiDAR point clouds in place of the pseudo point clouds generated from stereo images.
\begin{table}[htb]
 \caption{
 The $\rm AP_{3D}$ and $ \rm AP_{BEV}$ of ``Car'' on the \textbf{Lyft} validation set.
 }
 \small
 \centering
 \resizebox{\linewidth}{!}
 {
 \begin{tabular}{|c|ccc|ccc|}
  \hline
   \multirow{2}*{Method}      & \multicolumn{3}{c|}{$\rm AP_{3D}$ }     & \multicolumn{3}{c|}{$ \rm AP_{BEV}$}   \\
& Mod.      & Easy      & Hard      & Mod.      & Easy      & Hard \\
    \hline
    \hline
    LiDAR    & 59.01 & 80.01 & 47.19  & 66.21   & 88.18 &53.51 \\
    Mono & 20.76 & 33.41 &16.52  & 28.62   & 43.70 & 22.72 \\
    \hline
    MG-DA (Sec.~\ref{mg}) &24.28 &37.71 &18.76  &33.51 &50.42 &26.74 \\
    IoU-MA (Sec.~\ref{iou})    &21.58 &33.96 &17.39    &31.84 &47.91 &25.32 \\
    Mono $\rightarrow$ LiDAR & \textbf{25.55} & \textbf{39.46} & \textbf{19.70} & \textbf{33.74} & \textbf{50.61} &\textbf{26.98} \\
  \hline
\end{tabular}
}
\label{tab_lyft}
\end{table}

Table \ref{tab_lyft} shows that the proposed method outperforms the baseline on all three categories by a large margin in terms of both $\rm AP_{3D}$ and $\rm AP_{BEV}$.
It is obvious that significant performance gains can be achieved applying either the MG-DA or the IoU-MA individually.
By comparing the last row and the third row, we further find that adding the IoU-MA to the model with the MG-DA improves the AP values by 1.27 in the ``Mod.'' setting, indicating that the IoU-MA can compensate for inconsistent adaptation caused by mismatch.
This experiment on the Lyft dataset shows that adaptation from the monocular domain to the LiDAR domain at three levels aggressively forces the network to generate more discriminative features, which leads to better detection performance.
This experiment also demonstrates that our method can be applied to much larger datasets, thus validating the generalization ability of SGM3D.
\section{Conclusion}
In this paper, we introduce a novel stereo-guided monocular 3D object detection framework (SGM3D).
It generates pseudo-stereo representations from a monocular image by leveraging the proposed multi-granularity domain adaptation (MG-DA) mechanism and the IoU-matching-based alignment (IoU-MA) method.
Extensive experiments show that the proposed strategy owns state-of-the-art performance on the KITTI and Lyft datasets, and also potential practical values.

\bibliography{ieee_trans}
\bibliographystyle{IEEEtran}

\end{document}